\documentclass{article}
\usepackage{amssymb}

\usepackage{amsmath}
\usepackage{graphicx}
\usepackage{multirow}
\usepackage{url}
\usepackage{xcolor}
\usepackage{caption}
\usepackage{subcaption} 
\usepackage{float} 

\usepackage[final]{corl_2025} 

\title{\LARGE \bf
PicoPose: Progressive Pixel-to-Pixel Correspondence Learning for Novel Object Pose Estimation
}

\author{
  Liuhua Liu\textsuperscript{1}\quad
  Jiehong Lin\textsuperscript{2}\thanks{Corresponding author: mortimer.jh.lin@gmail.com.}\quad
  Zhenxin Liu\textsuperscript{1}\quad
  Kui Jia\textsuperscript{3}\\[0.5ex]
  \normalfont
  \textsuperscript{1}South China University of Technology \quad
  \textsuperscript{2}The University of Hong Kong \\
  \textsuperscript{3}The Chinese University of
Hong Kong, Shenzhen \\
}

\begin{document}
\maketitle

\begin{abstract}
RGB-based novel object pose estimation is critical for rapid deployment in robotic applications, yet zero-shot generalization remains a key challenge. In this paper, we introduce \textbf{PicoPose}, a novel framework designed to tackle this task using a three-stage pixel-to-pixel correspondence learning process. 
Firstly, PicoPose matches features from the RGB observation with those from rendered object templates, identifying the best-matched template and establishing coarse correspondences.
Secondly, PicoPose smooths the correspondences by globally regressing a 2D affine transformation, including in-plane rotation, scale, and 2D translation, from the coarse correspondence map.
Thirdly, PicoPose applies the affine transformation to the feature map of the best-matched template and learns correspondence offsets within local regions to achieve fine-grained correspondences.
By progressively refining the correspondences, PicoPose significantly improves the accuracy of object poses computed via PnP/RANSAC.
PicoPose achieves state-of-the-art performance on the seven core datasets of the BOP benchmark, demonstrating exceptional generalization to novel objects.
Code and trained models are available at \url{https://github.com/foollh/PicoPose}.


\end{abstract}              
\section{Introduction}

\begin{figure}[t]
  \centering
   \includegraphics[width=0.92\linewidth]{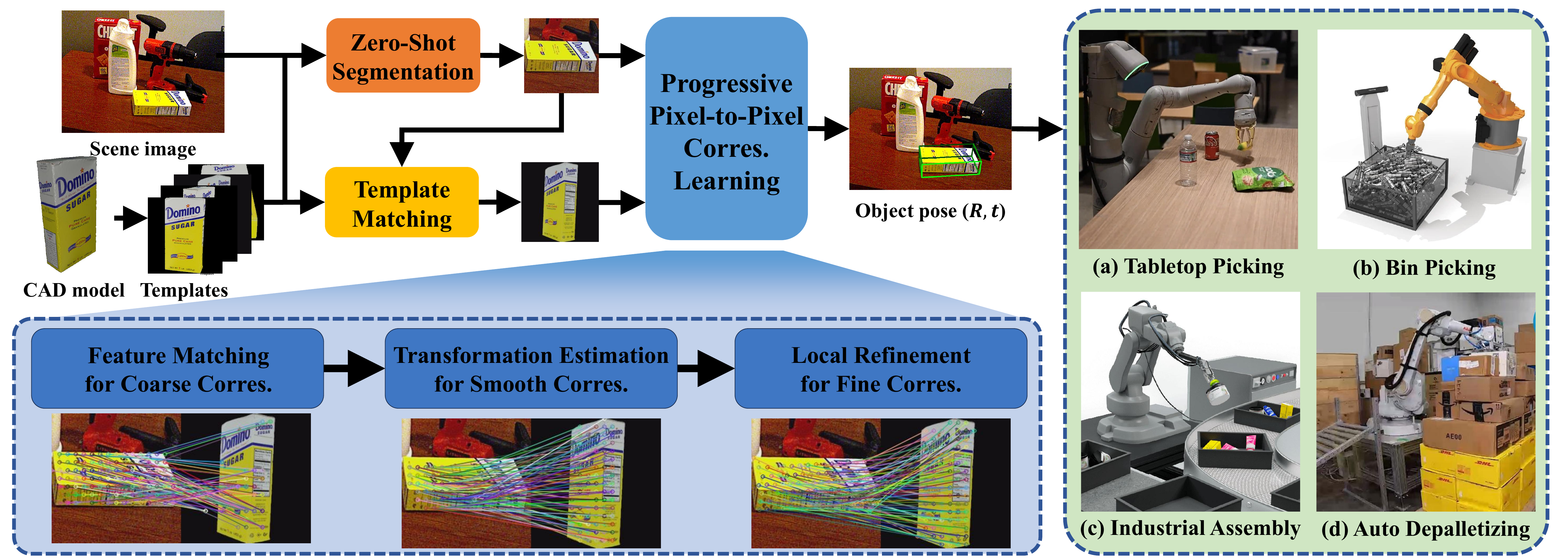}
   \caption{An overview of our proposed PicoPose with a three-stage pixel-to-pixel correspondence learning process for novel object pose estimation from RGB images. By progressively refining the correspondences, PicoPose significantly improves the accuracy of object poses computed via PnP/RANSAC. With zero-shot capability, PicoPose enables rapid deployment across various robotic manipulation systems for unseen objects.}
   \label{fig:head}
\end{figure}

Object poses are typically represented by six degrees of freedom (DoFs) parameters, including 3D rotation and translation, to define the transformation from a canonical object space to the camera space. Estimating object poses is highly sought after in real-world applications,  particularly in the context of robotics, where it enables precise manipulation, grasping, and interaction with various objects \cite{6907430, tremblay2018corl:dope, shridhar2022peract, Bin-Picking_Benchmark, RT-2} (see Fig. \ref{fig:head}), and is therefore extensively explored in research.

Early research \cite{xiang2017posecnn, peng2019pvnet, wang2021gdr, su2022zebrapose} primarily focused on pose estimation with the same object CAD models for both training and testing phases, but lacked flexibility for the objects unseen during training. 
Later studies \cite{wang2019normalized, lin2021dualposenet, lin2022category, di2022gpv, lin2023vi} addressed unseen objects within known categories by defining a normalized object coordinate space, but struggled with novel categories. 
With the advancement of foundation models \cite{radford2021learning, oquab2023dinov2, kirillov2023segment}, recent research \cite{labbe2022megapose, lin2024sam, wen2024foundationpose} has increasingly focused on handling new objects to achieve zero-shot pose estimation. While this capability enables rapid deployment of robotic systems, it presents significant generalization challenges that remain to be addressed.

For the zero-shot task of novel object pose estimation, recent methods using RGB-D images have achieved remarkable performance through techniques such as template matching with pose updating \cite{labbe2022megapose, wen2024foundationpose} or point registration for pose computation \cite{lin2024sam}. The success of these approaches is largely attributed to the essential geometric support provided by depth maps, which supply crucial features for matching and offer geometric priors that enhance object localization in 3D space.
However, the high cost of depth sensors often limits their practicality in real-world applications, making methods based solely on RGB images a more appealing option. Despite this, RGB-only approaches remain underexplored and generally fail to achieve competitive performance. 
Representative methods like GigaPose \cite{nguyen2024gigapose} and FoundPose \cite{ornek2025foundpose}, which rely on establishing correspondences between observed scenes and rendered templates via simply feature matching, often suffer from noisy correspondences prone to outliers, leading to imprecise pose predictions.

To this end,  we introduce a novel framework for progressive \underline{\textbf{pi}}xel-to-pixel \underline{\textbf{co}}rrespondence learning, termed as \textbf{PicoPose}, to enable precise pose estimation of novel objects from RGB images. As illustrated in Fig. \ref{fig:head}, PicoPose progressively refines the correspondences between RGB observations and templates across three stages, significantly enhancing the accuracy of object poses. With zero-shot capability, PicoPose enables rapid deployment across various robotic manipulation systems.

The architecture of PicoPose is illustrated in Fig. \ref{fig:overview}. More specifically, given an RGB image of a cluttered scene and a CAD model of an object that was not seen during training, PicoPose begins by rendering object templates from various viewpoints of the CAD model. These templates are then used in conjunction with zero-shot segmentation techniques (e.g., CNOS \cite{nguyen2023cnos}) to detect the target object within the RGB scene. PicoPose then uses a three-stage correspondence learning process to identify the best-matched template for the detected object and to learn fine-grained pixel-to-pixel correspondences between them. Since each pixel in the template corresponds to a 3D surface point on the CAD model, we establish pairs of 2D positions on the observation and the corresponding 3D points on the template, which are subsequently used to compute the 6D pose through PnP/RANSAC. For the process of correspondence learning, the three stages are described as follows:
\begin{itemize}
    \item \textbf{Stage 1: Feature Matching for Coarse Correspondences.} In this stage, PicoPose utilizes visual transformers to capture features for matching the RGB observation and the rendered templates, identifying the best-matched one and obtaining coarse correspondences.
    \item \textbf{Stage 2: Global Transformation Estimation for Smooth Correspondences.}  In this stage, PicoPose represents the coarse correspondences as a correspondence map, from which a 2D affine transformation, including in-plane rotation, scale, and 2D translation, is regressed to smooth the coarse correspondences and filter outliers.
    \item \textbf{Stage 3: Local Refinement for Fine Correspondences.} In this stage, PicoPose applies the affine transformation to the feature map of the best-matched template and employs several offset regression blocks to learn correspondence offsets within local regions.
\end{itemize}

We train PicoPose on the synthetic datasets of ShapeNet-Objects \cite{chang2015shapenet} and Google-Scanned-Objects \cite{downs2022google}, and test it on seven BOP datasets \cite{hodan2024bop}.
The quantitative results on these datasets outperform other existing methods, demonstrating the zero-shot capability of PicoPose. We also apply PicoPose in scenarios where object reference images are used to represent novel objects. 

In this paper, our key contributions are: (1) the introduction of PicoPose, a novel framework that leverages progressive pixel-to-pixel correspondence learning for pose estimation of novel objects from RGB images; (2) the development of three designed stages of correspondence learning to improve pose accuracy within PicoPose; and (3) the achievement of state-of-the-art results on the seven core datasets of the BOP benchmark for the RGB-based task, especially without refinement.

\section{Related Work}

\noindent \textbf{Methods Based on Image Matching.} To address the challenge of generalization, some approaches \cite{SA6D, cai2022ove6d, labbe2022megapose, nguyen2024gigapose, wen2024foundationpose, moon2024genflow} simplify the task of novel object pose estimation by using an image matching strategy, which involves rendering object templates in various poses and then retrieving the best-matched template to determine the corresponding pose. This strategy is often followed by downstream refinements, as in MegaPose \cite{labbe2022megapose} and GenFlow \cite{moon2024genflow}. In contrast, FoundationPose \cite{wen2024foundationpose} first updates the poses of templates before selecting the best-matched one.

\noindent \textbf{Methods Based on Pixel/Point Matching.} This group of methods estimates object poses by establishing correspondences, including 2D-3D correspondences for RGB inputs and 3D-3D correspondences for RGB-D inputs. For instance, OnePose \cite{sun2022onepose} matches the pixel descriptors in object proposals with the point descriptors obtained from Structure from Motion (SfM) to construct the 2D-3D correspondences, and OnePose++ \cite{he2022onepose++} further proposes coarse to fine matching to obtain more accurate correspondences. SAM-6D \cite{lin2024sam} learns 3D-3D correspondences through a two-stage point matching process incorporating background tokens. FoundPose \cite{ornek2025foundpose} leverages the generalization capabilities of foundation models to extract pixel features and establish 2D-3D correspondences.

\section{Method}
\subsection{Overview of PicoPose}

\begin{figure}[t]
  \centering
   \includegraphics[width=0.93\linewidth]{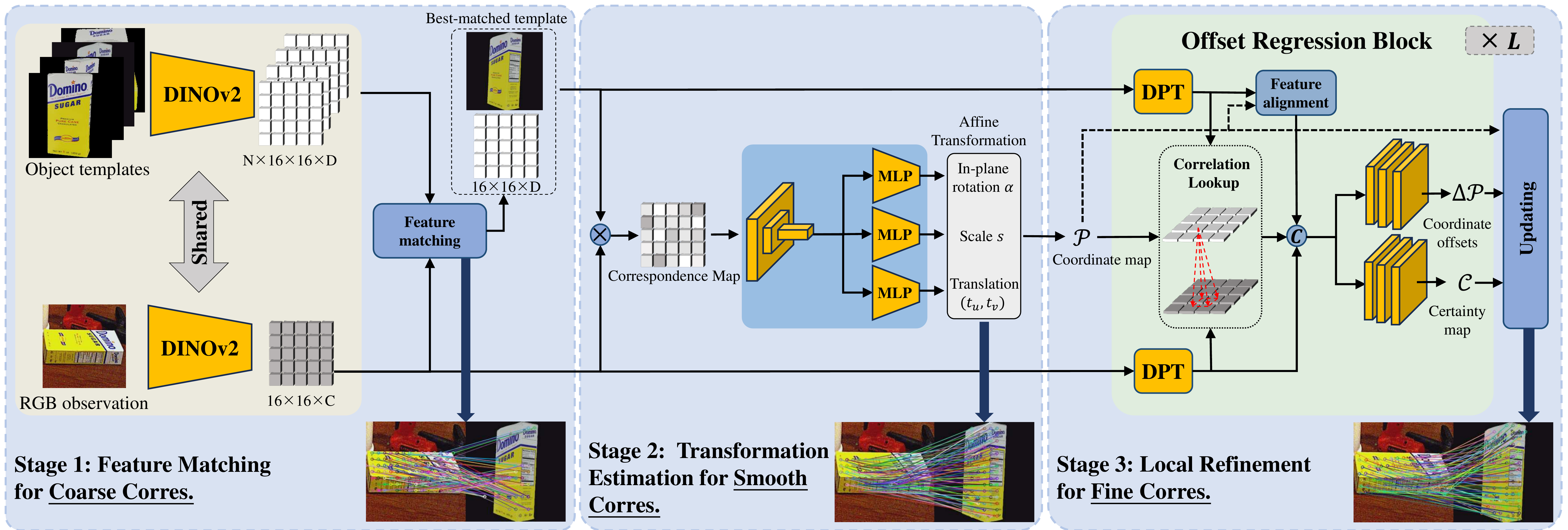}
   \caption{An illustration of our proposed \textbf{PicoPose}. 
   }
   \label{fig:overview}
\end{figure}

The goal of novel object pose estimation from RGB images is to determine the 6D transformation between an RGB observation and a CAD model of an object unseen during training. To address this task, we introduce a novel framework for generalizable \underline{\textbf{pi}}xel-to-pixel \underline{\textbf{co}}rrespondence learning, termed as \textbf{PicoPose}.
The architecture of PicoPose is illustrated in Fig. \ref{fig:overview}. For a given RGB image of a clustered scene and an object CAD model, we begin by rendering object templates from various viewpoints of the CAD model and employing zero-shot segmentation \cite{nguyen2023cnos} to identify and crop the region containing the target object from the RGB scene. We then resize both the detected crop and the object templates to a fixed size of $H \times W$, representing them as $\mathcal{I}$ and $\{\mathcal{T}_i\}_{i=1}^K$, respectively, where $K$ is the number of templates. PicoPose utilizes these inputs to search for the best-matched template, denoted as $\mathcal{T}$, and progressively learns the pixel-to-pixel correspondences between $\mathcal{I}$ and $\mathcal{T}$ across three stages, as detailed in Sec. \ref{subsec:method_progressive}. Each foreground pixel on $\mathcal{T}$ corresponds to a 3D surface point on the CAD model, enabling us to establish pairs of 2D positions on $\mathcal{I}$ and their corresponding 3D points on $\mathcal{T}$, which are utilized for computing the 6D object pose via PnP/RANSAC.

\subsection{Progressive Pixel-to-Pixel Correspondence Learning of PicoPose}
\label{subsec:method_progressive}

Given the resized detected crop $\mathcal{I} \in \mathbb{R}^{H\times W \times 3}$ and the object templates $\{\mathcal{T}_i \in \mathbb{R}^{H\times W \times 3}\}_{i=1}^K$
, PicoPose employs the ViT-L backbone \cite{dosovitskiy2020image}, pretranined by DINOv2 \cite{oquab2023dinov2}, to extract their patch features $\mathcal{F}_\mathcal{I} \in \mathbb{R}^{N\times D}$ and $\{\mathcal{F}_{\mathcal{T}_i} \in \mathbb{R}^{N\times D}\}_{i=1}^K$, respectively, where $D$ is the feature dimension and $N$ is the number of patches. PicoPose then identifies the best-matched templates $\mathcal{T}$ and performs three learning stages to build pixel-to-pixel correspondences between $\mathcal{I}$ and $\mathcal{T}$.

\noindent \textbf{Stage 1: Feature Matching for Coarse Correspondences}

Feature similarities between patches of the RGB observation $\mathcal{I}$  and the object templates $\{\mathcal{T}_i\}_{i=1}^K$, particularly with the best-matched template to $\mathcal{I}$, could establish coarse correspondences. To make it more effective, we initially retrieve the best-matched template $\mathcal{T}$ by scoring the degree of similarity between each template $\mathcal{T}_i$ and the observation $\mathcal{I}$. For each template $\mathcal{T}_i$, we obtain this template matching score $c_i$ by averaging the maximum feature cosine similarities of the foreground patches in $\mathcal{I}$ (identified by a prior-step zero-shot segmentation) with the patches in $\mathcal{T}_i$ as follows:
\begin{equation}
c_i = \frac{1}{N^{\prime}} \sum_{j\in\texttt{FG}(\mathcal{F}_\mathcal{I})} \max_{k=1,\dots,N} \frac{<\mathbf{f}_{\mathcal{I},j}, \mathbf{f}_{\mathcal{T}_{i}, k}>}{|\mathbf{f}_{\mathcal{I},j}| \cdot |\mathbf{f}_{\mathcal{T}_{i}, k}|},
\end{equation}
where $\mathbf{f}_{\mathcal{I},j} \in \mathcal{F}_\mathcal{I}$ and $\mathbf{f}_{\mathcal{T}_{i}, k} \in \mathcal{F}_{\mathcal{T}_i}$ are the $j^{th}$ patch features in $\mathcal{I}$ and the $k^{th}$ patch features in $\mathcal{T}_i$, respectively, with $<\cdot,\cdot>$ denoting an inner product. $\texttt{FG}(\mathcal{F}_\mathcal{I})$ represents the indices of foreground patches in $\mathcal{I}$, and $N^{\prime}$ is the count of foreground patches. From $\{\mathcal{T}_i\}_{i=1}^K$, the template $\mathcal{T}$ with the highest score is chosen as the best match. Furthermore, the feature similarities between $\mathcal{I}$ and $\mathcal{T}$ give coarse correspondences by determining the most similar patch in $\mathcal{T}$ for each patch in $\mathcal{I}$.

\noindent \textbf{Stage 2: Global Transformation Estimation for Smooth Correspondences}

In Stage 1, we exploit feature matching to obtain the coarse and sparse correspondences, which, however, often exhibit cluttered distributions with noise and many outliers. Therefore, the objective of this stage is to improve the smoothness of these correspondences and filter out the outliers.

To achieve the objective, we adopt a global approach to estimate the 2D affine transformation $\mathcal{M}$ between $\mathcal{I}$ and $\mathcal{T}$, which can be parameterized with 4 degrees of freedom (DoFs) \cite{nguyen2024gigapose} as follows:
\begin{equation}
\begin{aligned}
\mathcal{M} &= \begin{bmatrix}
    s \cos(\alpha) & -s \sin(\alpha) & t_{u} \\
    s \sin(\alpha) & s \cos(\alpha) & t_{v} \\
\end{bmatrix} \> ,
\end{aligned}
\label{eq:affine}
\end{equation}
where $\alpha$ denotes the in-plane rotation angle, $s$ denotes the relative scale between $\mathcal{I}$ and $\mathcal{T}$, and $(t_{u}, t_{v})$ represents the 2D translation of the object centroid in these two images. Applying $\mathcal{M}$ to transform the template $\mathcal{T}$ facilitates pixel alignments with $\mathcal{I}$, thus enabling smooth correspondences. 

\begin{figure}[t]
  \centering
   \includegraphics[width=0.86\linewidth]{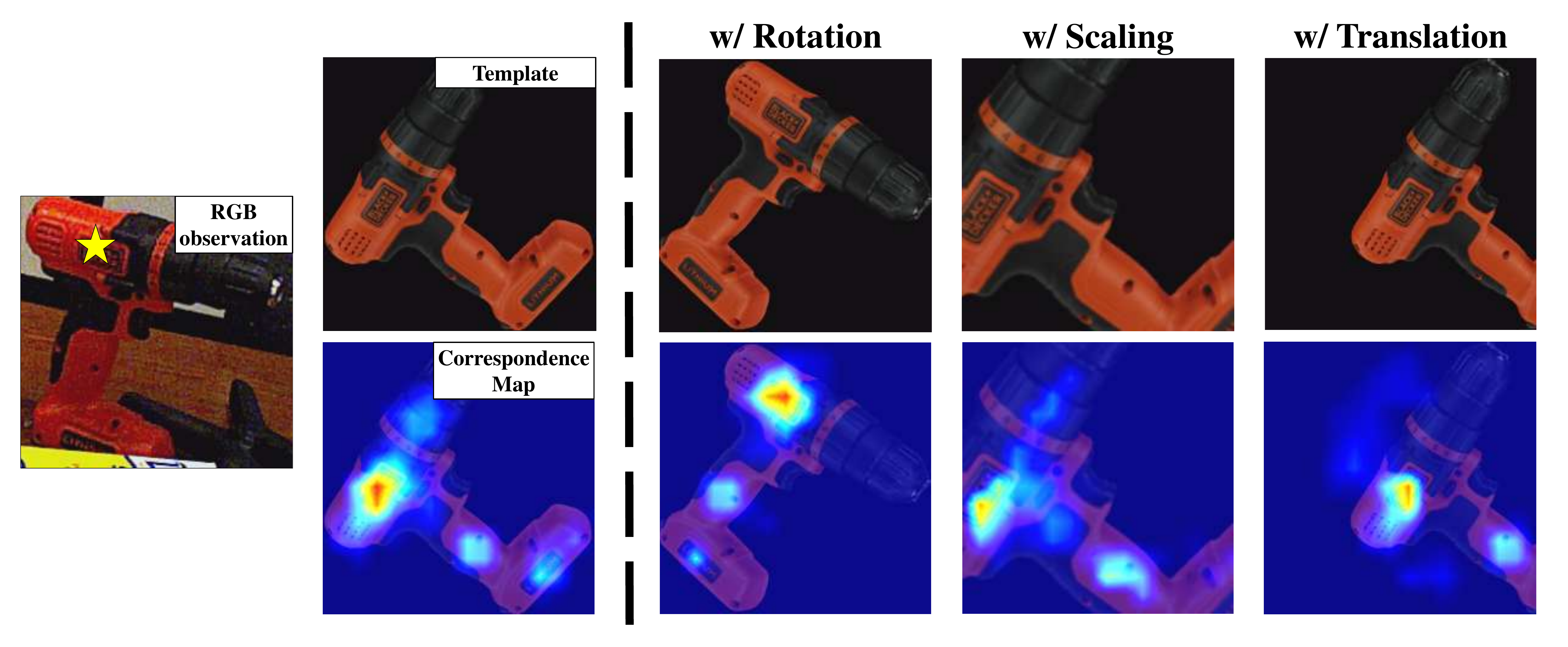}
   \vspace{-0.2cm}
   \caption{Visualization of correspondence maps between the feature of a point on the RGB observation (marked by a yellow star) and the features of templates with various affine transformations.
   }
   \vspace{-0.2cm}
   \label{fig:attention_map}
\end{figure}

As highlighted in Fig. \ref{fig:attention_map}, the correspondence map $\mathcal{A}$ between $\mathcal{I}$ and $\mathcal{T}$ can effectively capture the variations in $\alpha$, $s$, and $(t_{u}, t_{v})$, thereby encapsulating the essential patterns for learning $\mathcal{M}$. Therefore, 
instead of directly concatenating the patch features $\mathcal{F}_\mathcal{I}$ and $\mathcal{F}_\mathcal{T}$ for regressing $\mathcal{M}$, we propose a more effective approach by utilizing the coarse correspondences obtained in Stage 1, represented as the correspondence map $\mathcal{A}$ at this stage, to realize the target. More specifically, we first normalize the feature vectors of $\mathcal{F}_\mathcal{I}$ and $\mathcal{F}_\mathcal{T}$ to $\overline{\mathcal{F}}_\mathcal{I}$ and $\overline{\mathcal{F}}_\mathcal{T}$, respectively, and compute the correspondence map $\mathcal{A}$ as $\mathcal{A} = \overline{\mathcal{F}}_\mathcal{I}(\overline{\mathcal{F}}_\mathcal{T})^T \in \mathbb{R}^{N\times N} $, where $N=HW/196$. Subsequently, $\mathcal{A}$ is reshaped to the size of $(H/14) \times (W/14) \times N$ and passed through several stacked convolutions to reduce the spatial dimensions to a global pose vector. Finally, three parallel Multi-layer Perceptrons (MLPs) are applied to the pose vector to learn $(cos(\alpha), sin(\alpha))$, $s$ and $(t_{u}, t_{v})$ of $\mathcal{M}$, respectively.

\noindent \textbf{Stage 3: Local Refinement for Fine Correspondences}

With the transformation $\mathcal{M}$ predicted in Stage 2, we can align the feature map of the best-matched template $\mathcal{T}$ with the RGB observation $\mathcal{I}$, thereby achieving the smooth correspondences. For any position $(u, v)$ on $\mathcal{I}$, the corresponding position $(u^\prime, v^\prime)$ on $\mathcal{T}$ could be obtained using $\mathcal{M}$ as follows:
\begin{equation}
\begin{bmatrix}
 u^\prime\\
 v^\prime
\end{bmatrix} 
= \mathcal{M}
\begin{bmatrix}
 u\\
 v\\
 1
\end{bmatrix}.
\label{eqn:pos_trans}
\end{equation}
The deviation between $(u, v)$ and $(u^\prime, v^\prime)$ can be interpreted as the commonly known ``optical flow" \cite{xu2023unifying}. For all pixels in $\mathcal{I}$, we compute their corresponding positions in $\mathcal{T}$, denoted as $\mathcal{P}\in\mathbb{R}^{H\times W \times 2}$, via Eq. (\ref{eqn:pos_trans}) to represent the smooth correspondences. The affine transformation applied to the feature map of $\mathcal{T}$ is then achieved by using $\mathcal{P}$ as indices to gather features from $\mathcal{T}$, producing a transformed feature map aligned with $\mathcal{I}$. 
At this stage, we further learn the offsets $\Delta\mathcal{P}\in\mathbb{R}^{H\times W \times 2}$ to update $\mathcal{P}$ to $\mathcal{P}+\Delta\mathcal{P}$, enabling fine-grained correspondence adjustments within local regions.

We realize local refinement of correspondences in a progressive learning manner. Specifically, we first apply Dense Prediction Transformer (DPT) \cite{ranftl2021vision} to our backbone, generating $L$ hierarchical feature maps $\{\mathcal{F}_{\mathcal{I}l} \in \mathbb{R}^{H_l \times W_l \times D_l} \}_{l=1}^L$ and $\{\mathcal{F}_{\mathcal{T}l} \in \mathbb{R}^{H_l \times W_l \times D_l} \}_{l=1}^L$ of $\mathcal{I}$ and $\mathcal{T}$, respectively, where $H_l \times W_l$ denotes the spatial size and $D_l$ is the number of channels for the $l^{th}$ feature map. We then use $L$ offset regression blocks to iteratively update $\mathcal{P}$.

For the $l^{th}$ offset regression block, the current $\mathcal{P}$ is resized to $H_l \times W_l \times 2$ and scaled by dividing each 2D position within it by $[H/H_l$, $W/W_l]$ to ensure spatial consistency. We denote this resized and scaled version as $\mathcal{P}_l$, which we use as indices to gather features from $\mathcal{F}_{\mathcal{T}l}$, resulting in the transformed feature map $\mathcal{F}_{\mathcal{T}l}^{\prime}$ to align with $\mathcal{F}_{\mathcal{I}l}$. Additionally, we introduce a third correlation feature map  $\mathcal{F}_{\mathcal{C}l}$ via a Correlation Lookup module, introduced in RAFT \cite{teed2020raft}, to explicitly provide correlation degrees and facilitate easier learning of offsets; more details on this module can be found in the RAFT paper \cite{teed2020raft}. We then concatenate $\mathcal{F}_{\mathcal{I}l}$, $\mathcal{F}_{\mathcal{T}l}^{\prime}$, and $\mathcal{F}_{\mathcal{C}l}$ to form the input for two sequences of stacked convolutions, which are used to regress offsets $\Delta \mathcal{P}_l \in \mathbb{R}^{H_l \times W_l \times 2}$ and certainty map $\mathcal{S}_l \in \mathbb{R}^{H_l \times W_l}$. $\Delta \mathcal{P}_l$ is then interpolated to the size of $H \times W \times 2$, scaled by multiplying the 2D coordinates within it by $[H/H_l, W/W_l]$, and added to $\mathcal{P}$ for updating. The certainty map  $\Delta \mathcal{S}_l$ represents the confidence of the regressed offsets and is upsampled to generate $\mathcal{C}_l^\prime \in \mathbb{R}^{H \times W}$.

With $L$ offset regression blocks, we have the fine-grained $\mathcal{P}$, with a certainty map $\frac{1}{L} \sum_{i=1}^L \mathcal{C}_l^\prime$. For each foreground pixel in $\mathcal{I}$, if its correspondence certainty exceeds 0.5, we use the position in $\mathcal{P}$ to find the corresponding pixel in $\mathcal{T}$, which is linked to a 3D surface point. Therefore, all the pixel-to-pixel correspondences generate the associated 2D-3D pairs to compute the final object pose.

\subsection{Training of PicoPose}

We perform end-to-end training of the three-stage correspondence learning process in PicoPose by optimizing the following objective:
\begin{equation}
    \min \mathcal{L} = \mathcal{L}_{coarse} + \mathcal{L}_{smooth} + \mathcal{L}_{fine},
\end{equation}
where $\mathcal{L}_{coarse}$, $\mathcal{L}_{smooth}$, and $\mathcal{L}_{fine}$ are the loss terms associated with each of the three stages.

In Stage 1, we adopt the InfoNCE loss \cite{oord2018representation}, as used in GigaPose \cite{nguyen2024gigapose}, as the training objective $\mathcal{L}_{coarse}$ to learn feature matching.
In Stage 2, we predict the 2D affine transformation $\mathcal{M}$, including in-plane rotation angle $\alpha$, scale $s$, and 2D translation $(t_u, t_v)$, to generate smooth correspondences between $\mathcal{I}$ and $\mathcal{T}$. Letting $\hat{\alpha}$, $\hat{s}$, and $(\hat{t}_u, \hat{t}_v)$ represent the respective ground truths of the predicted parameters, we define the training objective $\mathcal{L}_{smooth}$ for this stage as follows:
\begin{equation}\label{loss:affine}
    \mathcal{L}_{smooth} = \mathcal{L}_{geo} (\alpha, \hat{\alpha}) + |\ln(s) - \ln(\hat{s})| + |t_u - \hat{t}_u| + |t_v - \hat{t}_v|,
\end{equation}
where $\mathcal{L}_{geo} (\alpha, \hat{\alpha})$ is the geodesic distance between two angles $\alpha$ and $\hat{\alpha}$, defined as follows:
\begin{equation}\label{loss:geodesic}
\begin{aligned}
    \mathcal{L}_{geo} (\alpha, \hat{\alpha}) =\text{acos}\bigl(\text{cos}(\alpha)\text{cos}(\hat{\alpha}) + 
    \text{sin}(\alpha)\text{sin}(\hat{\alpha})\bigr) \> .
\end{aligned}
\end{equation}
In Stage 3, we employ the $L_1$ distance and the binary cross entropy objective to guide the learning of coordinate offsets $\{\Delta \mathcal{P}_l\}_{l=1}^L$ and certainty maps $\{ \mathcal{C}_l\}_{l=1}^L$ across all $L$ offset regression blocks:
\begin{equation}
    \mathcal{L}_{fine} = \sum_{l=1}^L \lambda ||\hat{\mathcal{C}}_l \cdot (\Delta \mathcal{P}_l - \Delta \hat{\mathcal{P}}_l)|| +  \mu  \mathcal{L}_{bce} (\mathcal{C}_l, \hat{\mathcal{C}}_l),
    \label{eq:loss_fine}
\end{equation}
where $\{\Delta \hat{\mathcal{P}}_l\}_{l=1}^L$ and $\{ \hat{\mathcal{C}}_l\}_{l=1}^L$ represent the corresponding ground truths of $\{\Delta \mathcal{P}_l\}_{l=1}^L$ and  $\{ \mathcal{C}_l\}_{l=1}^L$, while $\lambda$ and $\mu$ are the weights to balance the loss terms. $\mathcal{L}_{bce}$ denotes the binary cross entropy objective. For the supervision of $\Delta \mathcal{P}_l$, we use $\hat{\mathcal{C}}_l $ to mask and exclude invalid correspondences.

\section{Experiments}

\begin{table*}
    \centering

    \resizebox{0.9 \textwidth}{!}{ 
    \begin{tabular}{c|c|ccccccc|c}
    \hline
    \multirow{2}{*}{Method}  & \multirow{2}{*}{$\#$Hypothesis} & \multicolumn{7}{c|}{BOP Dataset} & \multirow{2}{*}{Mean} \\
    \cline{3-9}
     & & LM-O & T-LESS & TUD-L & IC-BIN & ITODD & HB & YCB-V &\\
     \hline 
     \multicolumn{10}{c}{w/o Iterative Refinement} \\
    \hline
    MegaPose \cite{labbe2022megapose} & - & 22.9 & 17.7 & 25.8 & 15.2 & 10.8 & 25.1 & 28.1 & 20.8 \\
    GenFlow \cite{moon2024genflow}  & - & 25.0 & 21.5 & 30.0 & 16.8 & 15.4 & 28.3 & 27.7 & 23.5 \\
    GigaPose \cite{nguyen2024gigapose}  & 1  & 27.8 & 26.3 & 27.8 & 21.4 & 16.9 & 31.2 & 27.6 & 25.6 \\
    FoundPose \cite{ornek2025foundpose} & 1 & 39.5 & 39.6 & \textbf{56.7} & 28.3 & 26.2 & 58.5 & 49.7 & 42.6 \\
    PicoPose (Ours)   & 1 & \textbf{46.3} & \textbf{39.7} & 53.6 & \textbf{36.4} & \textbf{31.0} & \textbf{66.5} & \textbf{58.7} & \textbf{47.5} \\
    \hline
    GigaPose \cite{nguyen2024gigapose}   & 5 & 29.6 & 26.4 & 30.0 & 22.3 & 17.5 & 34.1 & 27.8 & 26.8 \\
    FoundPose w/o FM \cite{ornek2025foundpose}  & 5 & 39.6 & 33.8 & 46.7 & 23.9 & 20.4 & 50.8 & 45.2 & 37.2 \\
    FoundPose \cite{ornek2025foundpose}  & 5 & 42.0 &  \textbf{43.6} & \textbf{60.2} &  30.5 &  27.3 &  53.7 &  51.3  & 44.1 \\
    PicoPose (Ours)   & 5 & \textbf{49.2} & 41.3 & 58.4 & \textbf{37.8} & \textbf{32.7} & \textbf{67.6} & \textbf{57.6} & \textbf{49.2} \\ \hline 
     \multicolumn{10}{c}{w/ Refiner of MegaPose \cite{labbe2022megapose}} \\
    \hline
    MegaPose \cite{labbe2022megapose}   & 1 & 49.9 & 47.7 & \textbf{65.3} & 36.7 & 31.5 & 65.4 & 60.1 & 50.9 \\
    GigaPose \cite{nguyen2024gigapose}  & 1 & 55.7 & 54.1 & 58.0 & 45.0 & 37.6 & 69.3 & 63.2 & 54.7 \\
    FoundPose w/o FM \cite{ornek2025foundpose}   & 1 & 55.4 & 51.0 & 63.3 & 43.0 & 34.6 & 69.5 & 66.1 & 54.7 \\
    FoundPose \cite{ornek2025foundpose}   & 1 & 55.7 & 51.0 & 63.3 & 43.3 &  35.7 & 69.7 &  66.1 &  55.0 \\
    PicoPose (Ours)   & 1 & \textbf{60.5} & \textbf{56.6} & 63.6 & \textbf{46.5} & \textbf{40.1} & \textbf{75.9} & \textbf{68.7} & \textbf{58.8} \\
    \hline
    MegaPose \cite{labbe2022megapose}  & 5 & 56.0 & 50.7 & 68.4 & 41.4 & 33.8 & 70.4 & 62.1 & 54.7 \\
    GigaPose \cite{nguyen2024gigapose}   & 5 & 59.8 & 56.5 & 63.1 & 47.3 & 39.7 & 72.2 & 66.1 & 57.8 \\
    FoundPose w/o FM \cite{ornek2025foundpose}  & 5 & 58.6 & 54.9 & 65.7 & 44.4 & 36.1 & 70.3 & 67.3 & 56.8 \\
    FoundPose \cite{ornek2025foundpose}  & 5 & 61.0 & 57.0 & \textbf{69.4} &  47.9 & 40.7 & 72.3 & 69.0 & 59.6\\
    PicoPose (Ours) & 5 & \textbf{61.1} & \textbf{57.1} & 65.0 & \textbf{48.2} & \textbf{42.1} & \textbf{76.3} & \textbf{69.6} & \textbf{59.9} \\
    \hline
    \end{tabular}
    }
    \caption{Quantitative results of different methods on BOP datasets \cite{hodan2024bop}. We report the mean Average Recall (AR) among VSD, MSSD and MSPD. `FM' denotes featuremetric pose refinement \cite{ornek2025foundpose}.}
    \label{tab:result_sota}
\end{table*}

\noindent \textbf{Datasets.} we train PicoPose on synthetic datasets of ShapeNet-Objects \cite{chang2015shapenet} and Google-Scanned-Objects \cite{downs2022google} provided by \cite{labbe2022megapose}, using a total of 2 million training images. Evaluation is conducted on seven BOP datasets \cite{hodan2024bop}, including LM-O, T-LESS, TUD-L, IC-BIN, ITODD, HB, and YCB-V.

\noindent \textbf{Evaluation Metrics.} We report the mean Average Recall (AR)  w.r.t three error functions, i.e., Visible Surface Discrepancy (VSD), Maximum Symmetry-Aware Surface Distance (MSSD), and Maximum Symmetry-Aware Projection Distance (MSPD) \cite{hodan2024bop} . We also report the end-point-error (EPE), a widely used metric in flow estimation \cite{teed2020raft}, to assess the quality of the correspondences.

\begin{figure}[t]
  \centering
   \includegraphics[width=0.9\linewidth]{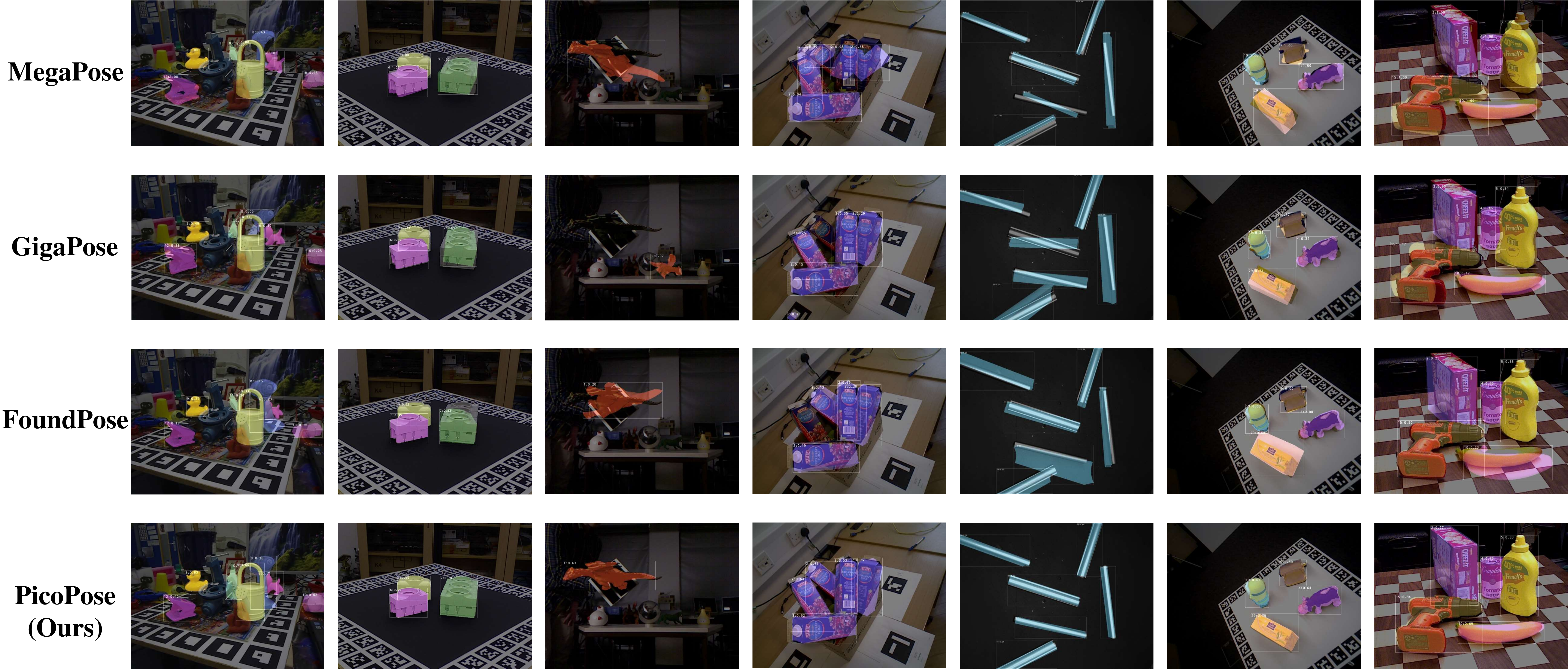}
   \caption{Qualitative results of different methods without iterative refinement on BOP datasets \cite{hodan2024bop}, including LM-O, T-LESS, TUD-L, IC-BIN, ITODD, HB, and YCB-V, arranged from left to right. }
   \label{fig:sota}
\end{figure}

\subsection{Comparisons with Existing Methods}
We evaluate PicoPose against existing methods on the seven core datasets of the BOP benchmark \cite{hodan2024bop}. Inspired by GigaPose \cite{nguyen2024gigapose}, we enhance the robustness of PicoPose by using the top 5 templates in Stage 2 and Stage 3 to learn fine correspondences and selecting the poses that best match these correspondences. The quantitative results are shown in Table \ref{tab:result_sota}, where PicoPose significantly outperforms other methods, highlighting its superior zero-shot capability for novel object pose estimation through progressive correspondence learning. For example, a single model of PicoPose using top 5 templates outperforms the single models of GigaPose \cite{nguyen2024gigapose} and FoundPose \cite{ornek2025foundpose} by $22.4\%$ and $5.1\%$ AR, respectively. In Table \ref{tab:result_sota}, we also report results with the iterative refinement proposed by MegaPose \cite{labbe2022megapose}, where PicoPose consistently outperforms the others,  whether using 1 or 5 pose hypotheses for refinement. The visualizations in Fig. \ref{fig:sota} further validate the advantages of PicoPose.

\begin{table}[t]
  \centering
  \begin{minipage}{0.48\linewidth}
    \centering
    \resizebox{0.8 \linewidth}{!}{
    \begin{tabular}{c|c|c}
    \hline
    Method & $\#$Hypothesis & AR \\ \hline\hline 
    GigaPose \cite{nguyen2024gigapose} & 1 & 17.9  \\
    PicoPose (Ours) & 1 & \textbf{20.4} \\
    \hline
    GigaPose \cite{nguyen2024gigapose} & 5 & 18.3 \\
    PicoPose (Ours) & 5 & \textbf{22.0} \\ \hline
    \end{tabular}}
    \vspace{0.1cm}
    \caption{Quantitative results of single reference image on LM-O. }
    \label{tab:result_single_image}

  \end{minipage}
  \hfill
  \begin{minipage}{0.48\linewidth}
    \centering
    \resizebox{0.98 \linewidth}{!}{
    \begin{tabular}{c|c|c|c}
    \hline
    Method & $\#$Hyp & Server & Time (s)      \\ \hline\hline
    GigaPose \cite{nguyen2024gigapose}  & 5 & NVIDIA V100 & 0.640    \\ \hline
    FoundPose \cite{ornek2025foundpose} & 5 & Tesla P100 & 3.360  \\ \hline
    GigaPose \cite{nguyen2024gigapose} & 1 & & 0.631 \\
    GigaPose  \cite{nguyen2024gigapose} & 5 & GeForce  & 1.209 \\
    PicoPose (Ours) & 1 &  RTX 3090 & 0.659  \\ 
    PicoPose (Ours) & 5 & &  1.562 \\ \hline 
    \end{tabular}}
    \vspace{0.1cm}
    \caption{Per-image runtime on LM-O. 
    }
    \label{tab:ablation_runtime}
  \end{minipage}
  \vspace{-0.5cm}
\end{table}

\noindent \textbf{Results with Single Object Reference Images.} Since obtaining perfect object CAD models is not always practical, object images are sometimes used as references. Following GigaPose \cite{nguyen2024gigapose}, we evaluate PicoPose in the most extreme scenario, i.e., with only one reference image, by employing Wonder3D \cite{long2024wonder3d} to reconstruct the object CAD model from this image. As shown in Table \ref{tab:result_single_image}, PicoPose can successfully handle this extreme setting, achieving results comparable to GigaPose.

\noindent \textbf{Runtime Analysis.} We report the per-image processing time, including segmentation and pose estimation, of different methods without iterative refinement in Table \ref{tab:ablation_runtime}. For a fair comparison, both GigaPose \cite{nguyen2024gigapose} and PicoPose are tested on the same servers; as shown in Table \ref{tab:ablation_runtime}, PicoPose achieves comparable speeds while delivering more impressive results, demonstrating its accuracy and efficiency. Notably, while FoundPose \cite{ornek2025foundpose} employs more advanced servers, it still incurs significantly higher computational costs when matching its extensive template library (800 templates per object). In contrast, both PicoPose and GigaPose use only 162 templates per object.

\begin{figure}[t]
  \begin{minipage}{0.44\linewidth}
    \centering
    \resizebox{0.95 \linewidth}{!}{
        \begin{tabular}{c|ccc|c}
        \hline
        Stage & LM-O & T-LESS & YCB-V  & MEAN          \\ \hline \hline
        \multicolumn{5}{c}{6D Pose Estimation (AR $\uparrow$)} \\ \hline
        1 & 28.6 & 27.3 & 41.2 & 32.4  \\ 
        2 & 31.2 & 25.6 & 45.5 &  34.1 \\ 
        3 & \textbf{46.3} & \textbf{39.7} & \textbf{58.7} & \textbf{48.2} \\
        \hline \hline 
        \multicolumn{5}{c}{Translation Estimation (Accuracy $\uparrow$)} \\ \hline
        1 & 40.3 & 43.1 & 60.2 & 47.9  \\ 
        2 & 43.2 & 48.4 & 69.6 & 53.7  \\
        3 & \textbf{62.6} & \textbf{56.0} & \textbf{78.7} & \textbf{65.8}  \\ 
        \hline \hline 
        \multicolumn{5}{c}{Correspondence Estimation (EPE $\downarrow$)} \\ \hline
        1     & 3.6 & 4.4 & 4.5 & 4.2  \\ 
        2     & 3.0 & 4.2 & 2.1 & 3.1  \\
        3     & \textbf{2.1}  &  \textbf{3.8}  &  \textbf{1.2}  & \textbf{2.4}   \\  \hline
        \end{tabular}}
        \captionof{table}{Quantitative comparisons among different stages of correspondence learning.}
        \label{tab:ablation_stage}
  \end{minipage}
  \hfill
  \centering
  \begin{minipage}{0.50\linewidth}
  \centering
   \includegraphics[width=0.98\linewidth]{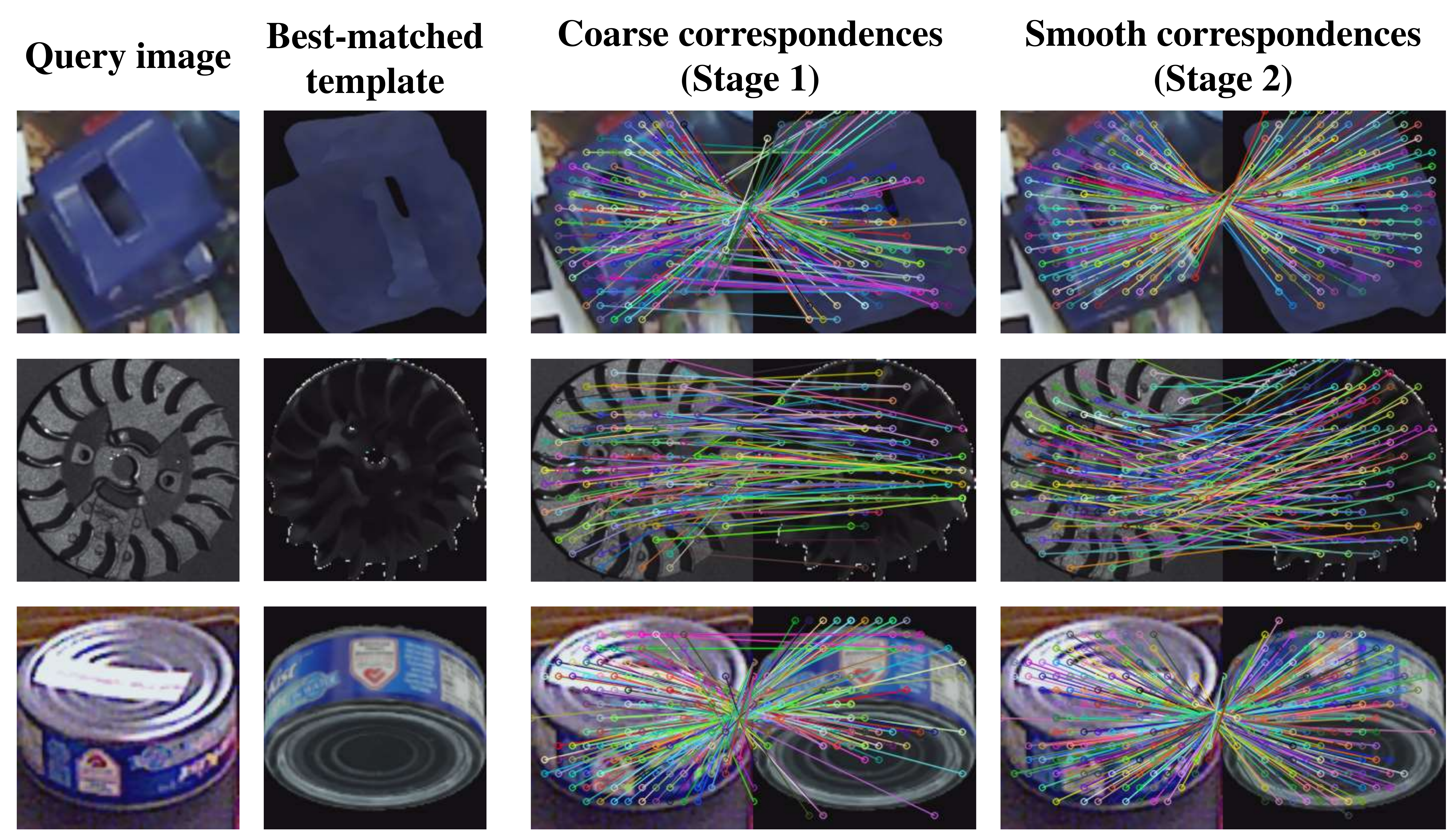}
   \caption{Visualization comparisons between the coarse correspondences from Stage 1 and the smooth ones from Stage 2. }
   \label{fig:corr}
  \end{minipage}
\end{figure}

\subsection{Ablation Studies and Analyses}

We conduct ablation studies to evaluate the efficacy of designs in PicoPose. Except for specific cases, the results are achieved using only the best-matched templates in Stage 2 and Stage 3.

\begin{table}[t]
\centering
\resizebox{0.9\textwidth}{!}{
\begin{tabular}{c|ccc|c|cc|c}
\hline
\multirow{2}{*}{Method}  & \multicolumn{4}{c|}{Pose Estimation (AR $\uparrow$)}  & \multicolumn{3}{c}{Processing Time on LM-O (s $\downarrow$)} \\ \cline{2-8}
& LM-O & T-LESS & YCB-V  & MEAN &  Model Forward & Post-Processing & ALL         \\ \hline \hline
$\mathbf{F}_{ist}$ in GigaPose \cite{nguyen2024gigapose} & 26.0 & 22.5 & 25.3 & 24.9 & 0.465 & 0.166 & 0.631 \\ \hline
Stage 2 w/ concatenated features  & 27.0  &  21.7  &  35.0  &  27.9  & 0.329 & - &  0.329       \\
Stage 2 w/ correspondence map  & \textbf{31.2}  &  \textbf{25.6}  &  \textbf{45.5}  &  \textbf{34.1}    & 0.329 &  - &  0.329      \\
\hline 
\end{tabular}
}
\vspace{0.1cm}
\caption{Quantitative comparisons among different variants of Stage 2.}
\vspace{-0.4cm}
\label{tab:ablation_corr_map}
\end{table}

\noindent \textbf{Efficacy of Progressive Correspondence Learning.} The key to the success of PicoPose lies in its design of progressive pixel-to-pixel correspondence learning. To evaluate this, we first analyze the quality improvements of correspondences and examine their impact on pose estimation. For Stage 2, the predicted affine transformations are applied to obtain correspondences. As shown in Table \ref{tab:ablation_stage}, pose precision improves with finer correspondences, supporting the core claim of this paper. Stage 2 smooths coarse correspondences and filters outliers (Fig. \ref{fig:corr}), significantly improving translation accuracy (errors $<5$ cm), while its overall 6D pose enhancement over Stage 1 is marginal, as Stage 2 does not include viewpoint rotation updates like Stage 1 with PnP. Stage 3 further enhances precision with a $14.1\%$ AR improvement by locally refining the correspondences from Stage 2.

\noindent \textbf{Efficacy of Stage 2.}  
We first assess the effectiveness of Stage 2 by presenting the results without it in Table \ref{tab:ablation_wo_stage2}, where we initialize the input correspondences for Stage 3 in two ways: 1) by directly using the coarse correspondences from Stage 1, and 2) by using the predicted poses from Stage 1 via PnP/RANSAC to obtain smoother correspondences. The first approach yields less precise results due to the high noise in the correspondences, while the second is less efficient because of the additional PnP/RANSAC processing. Next, we conduct experimental comparisons with regression based on direct feature concatenation of the observation and the template. As shown in Table \ref{tab:ablation_corr_map}, learning from correspondence maps proves to be more effective, as it explicitly models coarse correspondences and effectively captures the variations in affine transformations. Additionally, we replace our Stage 2 with $\mathbf{F}_{ist}$ from GigaPose \cite{nguyen2024gigapose}, which is less efficient, as discussed in Sec. \ref{subsec:method_progressive}.

\noindent \textbf{Efficacy of Stacked Offset Regression Blocks in Stage 3.} In Stage 3, we use $L$ offset regression blocks, specifically $L=3$ in our experiments with feature spatial sizes from DPT \cite{ranftl2021vision} set to $16 \times 16$, $32 \times 32$, and $64 \times 64$, to refine correspondences within local regions. The results from different offset regression blocks are reported in Table \ref{tab:ablation_offset}, where performance progressively improves as more blocks are used, indicating that finer correspondences are achieved through progressive local refinement.

\begin{table}[H]
   \centering
  \begin{minipage}[t]{0.48\textwidth}
    \resizebox{1.0\textwidth}{!}{
    \begin{tabular}{l|cc|cc}
    \hline
    \multirow{2}{*}{Initial Correspondence} & \multicolumn{2}{c|}{LM-O} & \multicolumn{2}{c}{T-LESS} \\  \cline{2-5}
    & AR & Time (s) & AR & Time (s) \\ \hline \hline
    Matching in Stage 1 & 34.5 & 0.655 & 25.0  &  0.612    \\
    Pose from Stage 1 & 45.7 & 0.936 &  33.9 &   0.823    \\ \hline
    Pose from Stage 2 & \textbf{46.3}  & 0.659 & \textbf{39.7} &  0.617\\
    \hline 
    \end{tabular}}
    \vspace{0.1cm}
    \captionof{table}{\small  Quantitative results with different initial correspondences inputted to Stage 3.}
    \label{tab:ablation_wo_stage2}
   \end{minipage}
  \hfill
  \begin{minipage}[t]{0.48\textwidth}
    \resizebox{1.0\textwidth}{!}{
    \begin{tabular}{c|c|ccc|c}
    \hline
    OR Block & Size & LM-O & T-LESS & YCB-V  & MEAN       \\ \hline \hline
    1  & $16\times 16$ &  33.4  &  27.2  &  42.3  &  34.3     \\
    2  & $32\times 32$ &  42.9  &  36.6  &  53.9  &  44.5     \\
    3  & $64\times 64$ &  \textbf{46.3}  &  \textbf{39.7}  &  \textbf{58.7}  &  \textbf{48.2}     \\ \hline 
    \end{tabular}}
    \vspace{0.1cm}
    \captionof{table}{\small Quantitative results of different offset regression blocks (denoted as ``OR Block") in Stage 3. 
    }
    \label{tab:ablation_offset}
  \end{minipage}
\end{table}

\section{Conclusion}

In this paper, we propose PicoPose, a novel framework for object pose estimation from RGB images that uses progressive pixel-to-pixel correspondence learning across three carefully designed stages. We demonstrate the zero-shot capabilities of PicoPose on seven core datasets of the BOP benchmark, enabling practical deployment in robotic applications. In future work, we aim to further increase the speed of PicoPose to achieve real-time performance and explore ways to reduce its reliance on templates.

\newpage
\section{Limitations}

While PicoPose demonstrates strong performance on 6D object pose estimation benchmarks with real-world cluttered scenes and shows promising potential for rapid deployment in robotic applications through simulated grasping experiments, several areas remain for future improvement. 

First, the reliance on multiple templates for 3D object representation means that its performance is inherently tied to the number of templates used. Although our approach achieves comparable results with fewer templates than existing methods (as detailed in Supplementary Section 2), future work could explore more efficient template utilization strategies or alternative 3D representations. 

Second, while PicoPose benefits from the iterative refinement of MegaPose \cite{labbe2022megapose}, the performance gains are relatively modest compared to other methods, probably because PicoPose's initial estimates already approach the refiner's performance upper bound. This motivates the development of specialized refinement approaches for high-quality initial predictions. 

Third, in case of a single reference image of the target object, where we follow GigaPose \cite{nguyen2024gigapose} to use Wonder3D for reconstructing the object in 3D space, the quality of reconstruction remains a limiting factor that affects final pose estimation accuracy.  We believe integrating emerging neural reconstruction techniques could significantly improve performance in this challenging scenario.

\bibliography{example}

\newpage
\appendix
\section*{Appendix}
This appendix is structured as follows. 
In \textbf{Section \ref{sec: more-experimental-details}}, we provide more details on the network architecture, training setup, equipment, pose solving configuration, and data augmentation methods used in this work.
In \textbf{Section \ref{sec: more-ablation-study}}, we perform more ablation studies on the effects of correlation lookup in Stage 3 and the influence of the number of templates on pose precision.
\textbf{Section \ref{sec: more-qualitative-results}} shows more qualitative results of different stages in PicoPose and more qualitative comparisons of different methods on real-world benchmarks.
In \textbf{Section \ref{sec: robot-simulation-grasping-experiment}}, we conduct robotic grasping experiments in a simulation environment using the PyBullet physics engine \cite{pybullet} to verify the application of our method in robotic grasping.

\section{More Experimental Details}
\label{sec: more-experimental-details}
\textbf{Network architecture details.} 
In Stage 1, we build on previous work \cite{nguyen2024gigapose} by using dinov2$\_$vitl14 as our backbone. The feature dimension of this backbone is $1,024$ for each token.
In Stage 2, we employ two conventional layers with group normalization and a ReLU activation function to reduce the spatial size of the input correspondence map $\mathcal{A}$ to $8 \times 8$, after which we flatten the feature map to obtain the global pose vector.
In Stage 3, in order to reduce network parameters, we set the number of channels $D_l$ of the $l^{th}$ feature map generated from DPT \cite{ranftl2021vision} to 256. Additionally, we utilize the standard lookup operation in RAFT \cite{teed2020raft} for $L$ blocks. Since the feature sizes in each block vary, the hyperparameter settings of each block need to be adjusted individually. For the $l^{th}$ block, we establish the layer of the correlation pyramid as $l+1$ and set the radius of the correlation lookup to 4.
We also list the model sizes of different stages in Table \ref{tab:model_size}.

\begin{table}[ht]
    \centering
    \begin{tabular}{c|c}
    \hline
    Stage & \#Param  \\
    \hline
    1 & 304 M \\ 
    2 & 18 M \\
    3 & 58 M  \\
    \hline
    Total & 380 M \\
    \hline
    \end{tabular}
    \caption{The model sizes of different stages in our network.}
    \label{tab:model_size}
\end{table}

\textbf{Details of training settings.} 
We report the detailed hyperparameter settings to train our network in Table \ref{tab:hyperparameter}.

\begin{table}[ht]
    \centering
    \begin{tabular}{c|c}
    \hline
    Hyperparameters & Settings  \\
    \hline
    Optimizer & AdamW \\ 
    AdamW $\beta$ & (0.5, 0.99) \\
    AdamW $eps$  & 1e-6  \\
    Learning rate scheduler &  Cosine decay \\ 
    Training iterations & $400,000$ \\ 
    Warmup iterations & $1,000$  \\ 
    Learning rate & 1e-5 \\
    Weight decay & 5e-4 \\
    Batch size & 32 \\
    \hline
    \end{tabular}
    \caption{Detailed hyperparameters in training our network. }
    \label{tab:hyperparameter}
\end{table}

\textbf{More details of devices.}
We conducted all experiments with GPU in GeForce RTX 3090 24G, and CPU in Intel (R) Xeon (R) CPU E5-2678 v3 @ 2.50 GHz under the Linux operating system.

\textbf{Details of PnP/RANSAC.}
We utilize the EPnP algorithm \cite{lepetit2009ep} along with the RANSAC scheme in the fine correspondence to solve the object pose in Stage 3. The RANSAC iterations are configured to 150, and the reprojection error threshold is set to 2.

\textbf{Details of augmentations.} 
In Stage 1, to better adapt to the input images in real-world scenarios, we conduct data augmentation on the query image of the training data in a manner similar to GDRNPP \cite{liu2022gdrnpp_bop}. In Stage 3, we apply random 2D translation, in-plane rotation, and scale noise to the ground truth affine transformation $\mathcal{M}$ to generate the initial coordinate map $\mathcal{P}$.

\section{More Ablation Studies}
\label{sec: more-ablation-study}
\textbf{Effects of correlation lookup in Stage 3.} 
In Stage 3, we use the correlation lookup operation in RAFT \cite{teed2020raft} to obtain flow features, and combine them with the features of the input image and the best-matched template to predict coordinate offsets and the certainty map. To verify the effectiveness, we conducted experiments without using the correlation lookup operation. As shown in Table \ref{tab:ablation_stage3}, the features obtained by the correlation lookup operation can improve the results significantly.

\begin{table}[ht]
\centering
\begin{tabular}{c|ccc|c}
\hline
Correlation Lookup & LM-O & T-LESS & YCB-V  & MEAN          \\ \hline
$\times$ & 35.0 & 25.9 & 46.8 & 35.9  \\ 
\checkmark & \textbf{46.3} & \textbf{39.7} & \textbf{58.7} & \textbf{48.2}        \\ \hline
\end{tabular}
\caption{Quantitative results of correlation lookup operation in Stage 3. We report the mean Average Recall (AR) among VSD, MSSD and MSPD. 
}
\label{tab:ablation_stage3}
\end{table}

\textbf{Influence of the number of templates.}
We follow the setup of GigaPose \cite{nguyen2024gigapose} by using $K=162$ templates per object in our evaluation experiments. In Table \ref{tab:ablation_template}, we present additional quantitative results with different numbers of templates for both GigaPose and our proposed PicoPose. The results show improvement as more templates are used, since both methods rely on template matching to select the best-matched template for the target object. However, the rate of improvement slows as the number of templates increases. Notably, PicoPose is more effective than GigaPose when using fewer templates, further highlighting the advantages of PicoPose.

\begin{table}[ht]
\centering
\begin{tabular}{c|c|ccc|c}
\hline
Method & $\#$Temp & LM-O & T-LESS & YCB-V  & MEAN          \\ \hline \hline
GigaPose \cite{nguyen2024gigapose} & \multirow{2}{*}{2} & 4.8 &  4.4  & 2.0  & 3.7 \\
PicoPose & & \textbf{10.5} & \textbf{12.8} & \textbf{14.1} & \textbf{12.5} \\ \hline
GigaPose \cite{nguyen2024gigapose} & \multirow{2}{*}{6} & 11.5 & 9.2 & 5.9 & 8.9 \\
PicoPose & & \textbf{27.5} & \textbf{25.0} & \textbf{39.5} & \textbf{30.7} \\ \hline
GigaPose \cite{nguyen2024gigapose} & \multirow{2}{*}{42} & 25.0 & 23.3 & 23.4 & 23.9 \\
PicoPose& &  \textbf{43.9} & \textbf{37.9} & \textbf{57.4} & \textbf{46.4}  \\ \hline
GigaPose \cite{nguyen2024gigapose} & \multirow{2}{*}{162} & 29.6 & 26.4 & 27.8 & 27.9 \\ 
PicoPose & & \textbf{46.3} & \textbf{39.7} & \textbf{58.7} & \textbf{48.2}  \\ \hline
\end{tabular}
\caption{Quantitative comparison with GigaPose \cite{nguyen2024gigapose} on the number of templates. We report the mean Average Recall (AR) among VSD, MSSD and MSPD.}
\label{tab:ablation_template}
\end{table}

\section{Additional Qualitative Results}
\label{sec: more-qualitative-results}
\begin{figure}[t]
  \centering
   \includegraphics[width=0.98\linewidth]{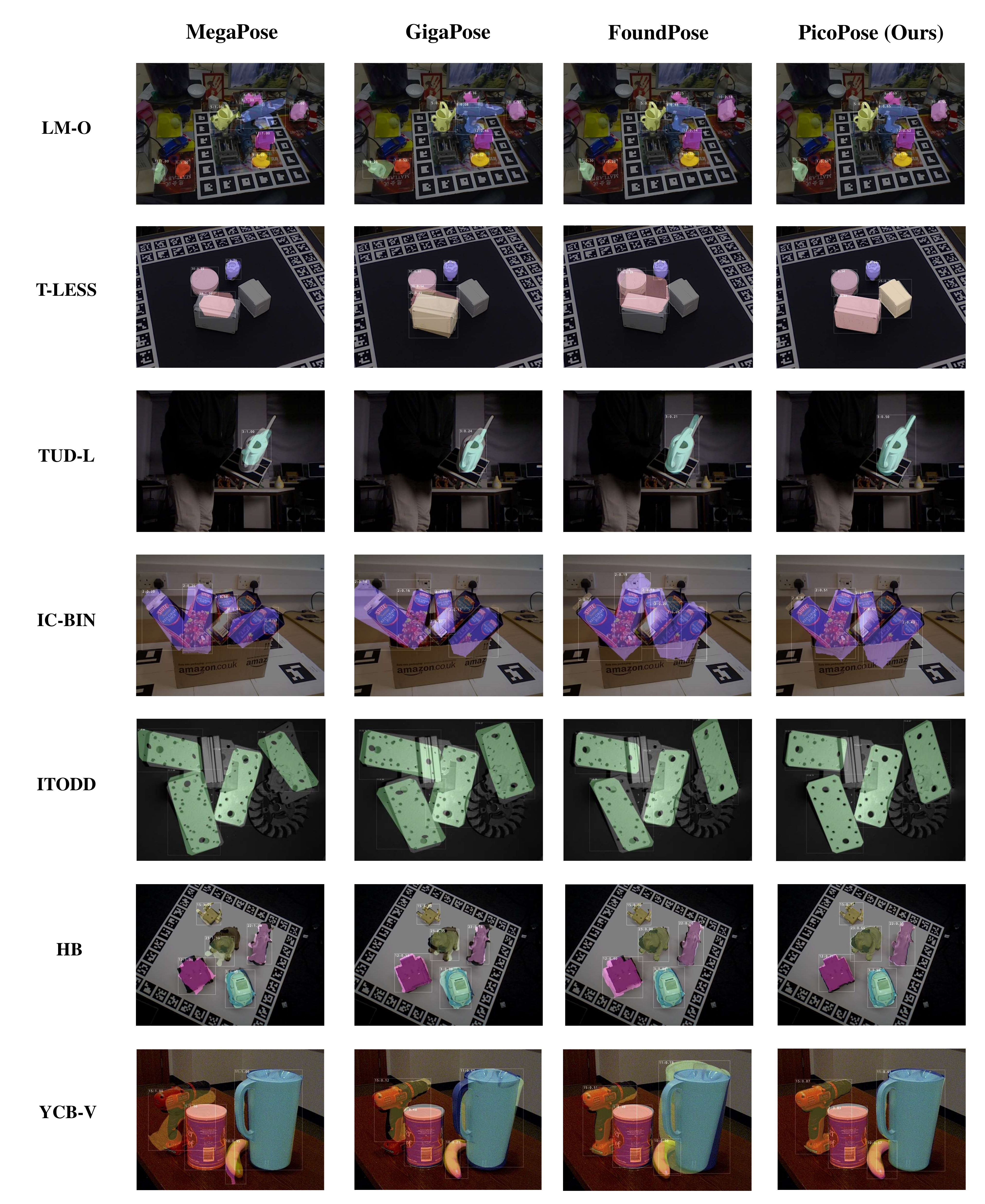}
   \caption{More qualitative results of different methods on the seven core datasets of BOP benchmark \cite{hodan2024bop}, including LM-O, T-LESS, TUD-L, IC-BIN, ITODD, HB, and YCB-V, arranged from top to bottom.}
   \label{fig:vis_pose_supp}
\end{figure}

\textbf{More qualitative results of different methods.}
We present more qualitative results of different methods on the seven core datasets 
(LM-O\cite{brachmann2014learning}, T-LESS\cite{hodan2017tless}, TUD-L\cite{hodan2018bop}, IC-BIN\cite{doumanoglou2016recovering}, ITODD\cite{drost2017introducing}, HB\cite{kaskman2019homebreweddb}, and YCB-V\cite{xiang2017posecnn} )
in the BOP benchmark \cite{hodan2024bop}, shown in Fig. \ref{fig:vis_pose_supp}. We illustrate the estimated 6D pose by rendering the 3D model on the input image and using the overlap rate as a basis, where a higher overlap rate indicates a more accurate estimated 6D pose. Specifically, all methods use the same zero-shot segmentation method, i.e., CNOS \cite{nguyen2023cnos}.

\begin{figure}
  \centering
   \includegraphics[width=0.98\linewidth]{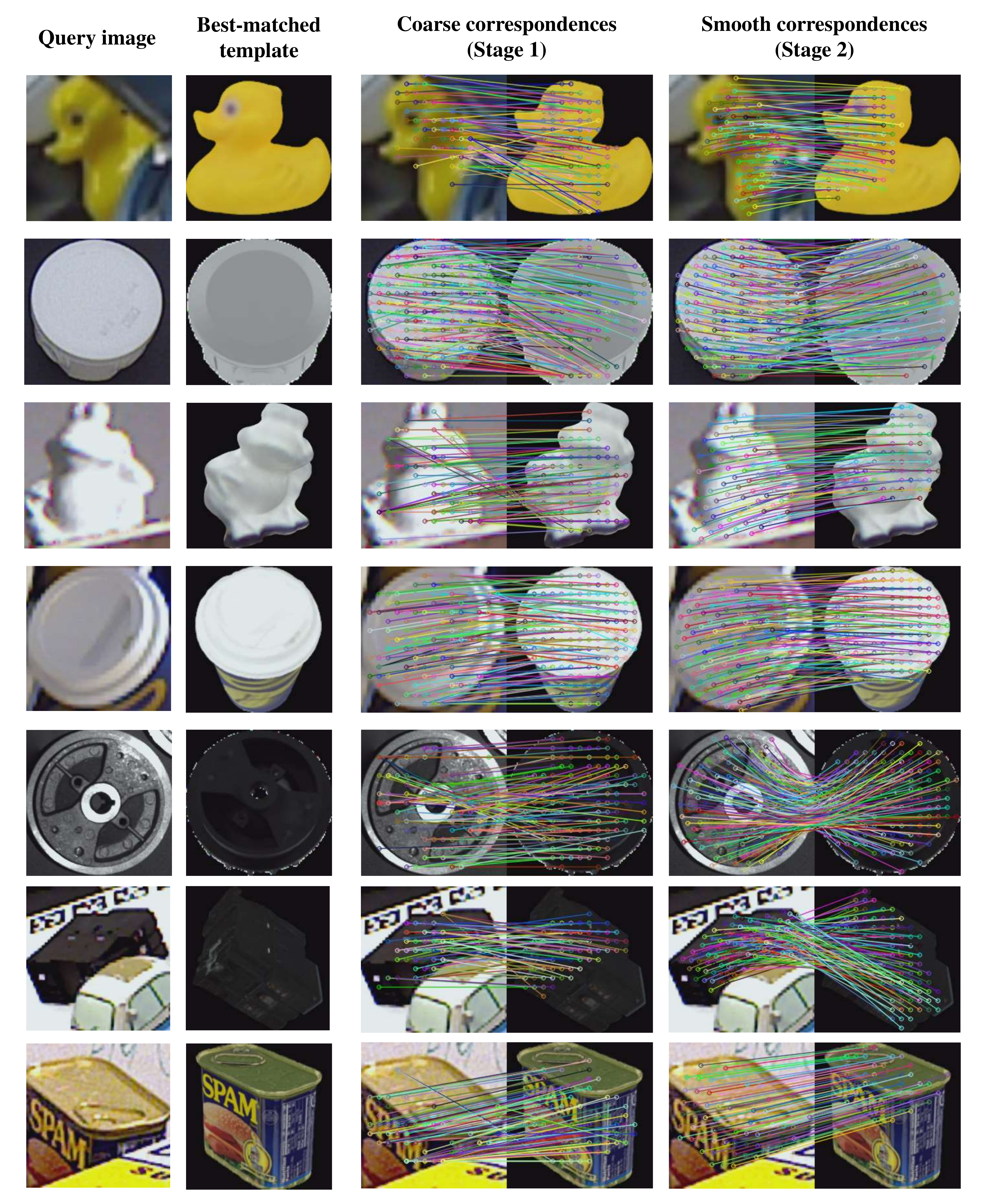}
   \caption{Qualitative results of coarse correspondences in Stage 1 and smooth correspondences in Stage 2 on the seven core datasets of BOP benchmark \cite{hodan2024bop}, including LM-O, T-LESS, TUD-L, IC-BIN, ITODD, HB, and YCB-V, arranged from top to bottom.}
   \label{fig:vis_stage_1_2}
\end{figure}

\begin{figure}
  \centering
   \includegraphics[width=0.98\linewidth]{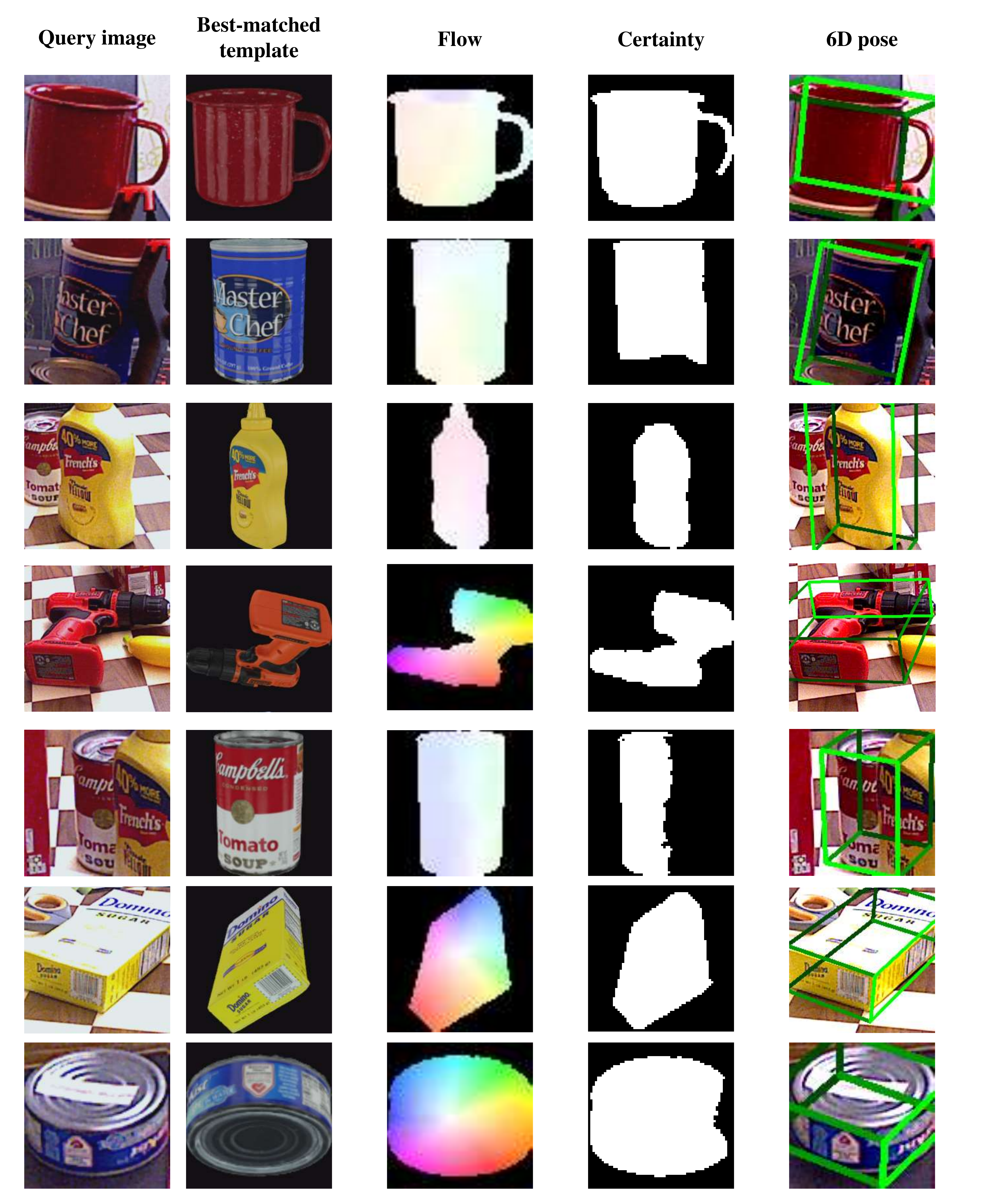}
   \caption{Qualitative results of fine correspondences in Stage 3 on YCB-V dataset \cite{xiang2017posecnn}.}
   \label{fig:vis_stage_3}
\end{figure}

\textbf{More qualitative results of different stages.}
We visualize the correspondences between the query image $\mathcal{I}$ and the best-matched template $\mathcal{T}$ in Stage 1 and Stage 2. As shown in Fig. \ref{fig:vis_stage_1_2}, the coarse correspondences generated in Stage 1 contain numerous outliers and inconsistencies, many of which are effectively resolved by Stage 2 to produce smooth correspondences. In Stage 3, we enhance the display of fine correspondences by visualizing the coordinate map $\mathcal{P}$ as optical flow with the certainty map on YCB-V dataset \cite{xiang2017posecnn}, shown in Fig. \ref{fig:vis_stage_3}.

\newpage
\section{Application: Robotic Grasping in Simulated Environments}
\label{sec: robot-simulation-grasping-experiment}

\begin{figure}[ht]
  \centering
   \includegraphics[width=0.98\linewidth]{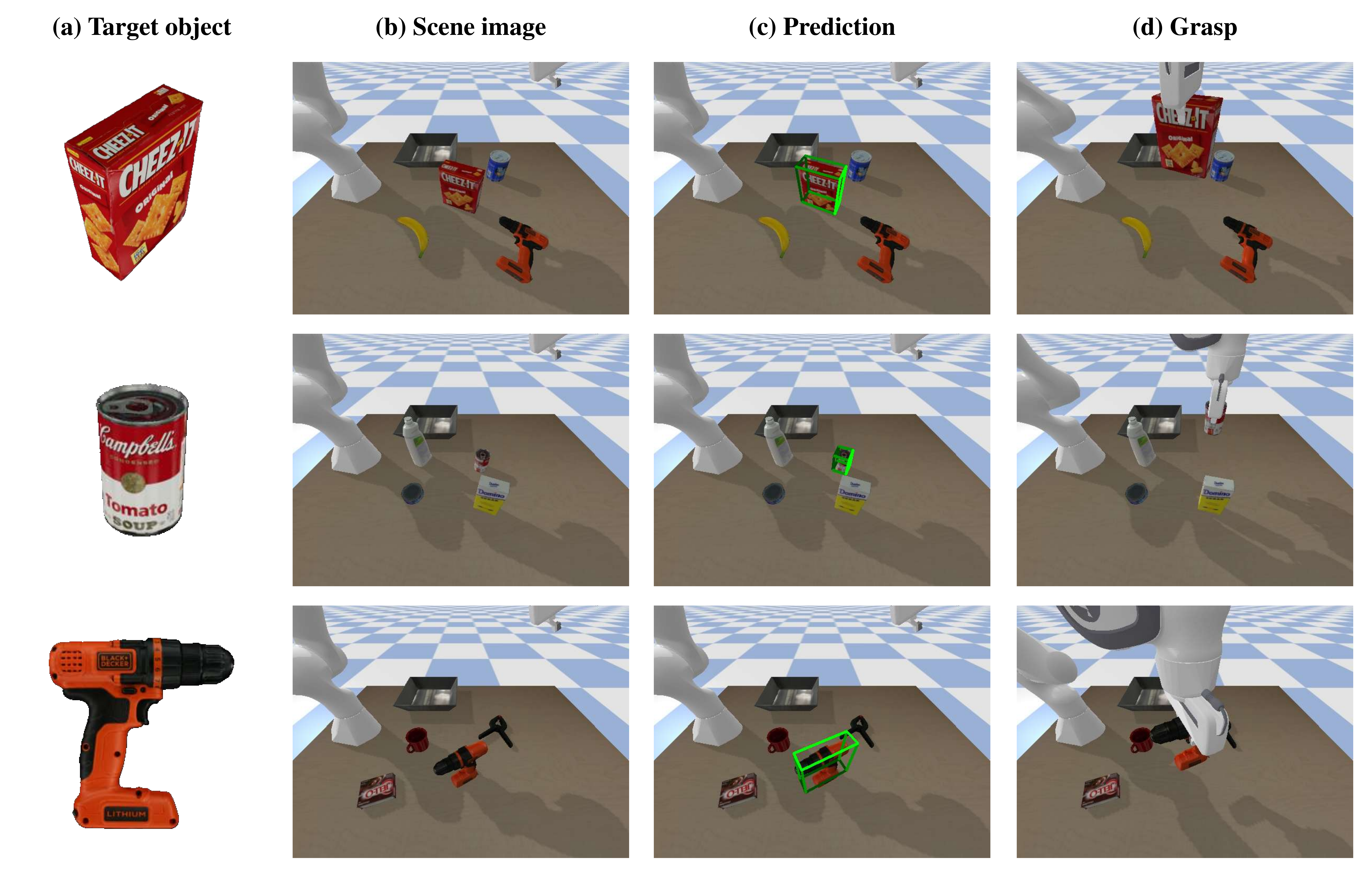}
   \caption{
   Robotic grasping application of PicoPose in simulated environment.
   }
   \label{fig:robot_grasp}
\end{figure}

In this section, our proposed PicoPose demonstrates seamless integration for robotic grasping applications using PyBullet \cite{pybullet} with the setup shown in Fig. \ref{fig:robot_grasp}. Our experimental scene comprises (1) a Franka Emika Panda robotic arm, (2) distractor objects, including the target, randomly arranged on the workspace, and (3) a placement tray. A fixed virtual camera captures single RGB images of the cluttered scene as input to our system.

The processing pipeline consists of three key stages. First, CNOS \cite{nguyen2023cnos} segments the target object in the RGB scene. Second, our proposed PicoPose estimates the 6D pose of the target object in camera coordinates. Third, we use the known camera-to-robot coordinate transformation to convert this pose into a 6D grasping pose and use inverse kinematics to generate robot motions to successfully grasp the target object.

\end{document}